\documentclass{article}
\usepackage{ijcai23}

\usepackage{times}
\usepackage{soul}
\usepackage{url}
\usepackage[hidelinks]{hyperref}
\usepackage[utf8]{inputenc}
\usepackage[small]{caption}
\usepackage{graphicx}
\usepackage{amsmath}
\usepackage{amsthm}
\usepackage{booktabs}
\usepackage{algorithm}
\usepackage{algorithmic}
\usepackage[switch]{lineno}
\usepackage{amssymb}
\usepackage{amsmath}
\usepackage{booktabs,tabularx}
\usepackage{multirow,comment,xcolor}
\usepackage{rotating}

\title{EvaNet: Elevation-Guided Flood Extent Mapping on Earth Imagery (Extended Version)}

\author{
Mirza Tanzim Sami$^2$
\and
Da Yan$^1$\and
Saugat Adhikari$^1$\and
Lyuheng Yuan$^1$\\
Jiao Han$^1$\and
Zhe Jiang$^3$\and
Jalal Khalil$^4$\And
Yang Zhou$^5$
\affiliations
$^1$Indiana University Bloomington, $^2$The University of Alabama at Birmingham\\
$^3$University of Florida, $^4$St.\ Cloud State University, $^5$Auburn University\\
\emails
mtsami@uab.edu,
\{yanda,adhiksa,lyyuan,jiaohan\}@iu.edu,
zhe.jiang@ufl.edu,
jalal.khalil@stcloudstate.edu,
yangzhou@auburn.edu
}

\begin{document}

\maketitle

\begin{abstract}
Accurate and timely mapping of flood extent from high-resolution satellite imagery plays a crucial role in disaster management such as damage assessment and relief activities. However, current state-of-the-art solutions are based on U-Net, which cannot segment the flood pixels accurately due to the ambiguous pixels (e.g., tree canopies, clouds) that prevent a direct judgement from only the spectral features. Thanks to the digital elevation model (DEM) data readily available from sources such as United States Geological Survey (USGS), this work explores the use of an elevation map to improve flood extent mapping. We propose, EvaNet, an elevation-guided segmentation model based on the encoder-decoder architecture with two novel techniques: (1)~a loss function encoding the physical law of gravity that {\em if a location is flooded (resp.\ dry), then its adjacent locations with a lower (resp.\ higher) elevation must also be flooded (resp.\ dry)}; (2)~a new (de)convolution operation that integrates the elevation map by a location-sensitive gating mechanism to regulate how much spectral features flow through adjacent layers. Extensive experiments show that EvaNet significantly outperforms the U-Net baselines, and works as a perfect drop-in replacement for U-Net in existing solutions to flood extent mapping. EvaNet is open-sourced at \url{https://github.com/MTSami/EvaNet}.
\end{abstract}

\vspace{-2mm}
\section{Introduction}
Climate change is drastically increasing the intensity and occurrence of floods~\cite{matgen2020feasibility}. In just the last two decades, flooding has negatively impacted over 2.3 billion people, often disproportionately affecting vulnerable communities ~\cite{wahlstrom2015human}. 
Therefore, accurate and timely mapping of flood extent can be crucial for effectively planning rescue and rehabilitation efforts ~\cite{oddo2019value}. 
However, sending field crews on site for inspection is prohibitively slow, while obtaining data from a limited number of groundwater sensors or UAVs cannot provide a global coverage.
Therefore, we target the accurate and timely mapping of flood extent from high-resolution satellite imagery that provides a global coverage.

There has been a lot of research on using remote-sensing satellite imagery for flood extent mapping. Two commonly used satellite types are radar (active sensor) and optical (passive sensor). For example, Sentinel-1~\cite{Sentinel93:online} and Sentinel-2~\cite{Sentinel11:online} are radar and optical satellites, respectively, that produce high resolution earth imagery at high revisit times~\cite{matgen2020feasibility}. 
Radar is more resilient to noises like clouds, but traditional approach to generating flood maps from radar imagery relies on semi-automated tools,~\cite{rudner2019multi3net} or thresholding techniques~\cite{fraccaro2022deploying} that are slow and lack generalizability, often requiring expert intervention for parameter tuning. 
In this paper, we focus on the optical satellite imagery, since radar data such as that from Sentinel-1~\cite{Sentinel93:online} can also be converted into RGB channels~\cite{fraccaro2022deploying}. 
Moreover, both \cite{konapala2021exploring} and \cite{bonafilia2020sen1floods11} found that using Sentinel-2 data only yields the best segmentation performance for deep learning models, as compared with when Sentinel-1 data is involved.

\begin{figure}[t]
    \centering
    \includegraphics[width=0.84\columnwidth]{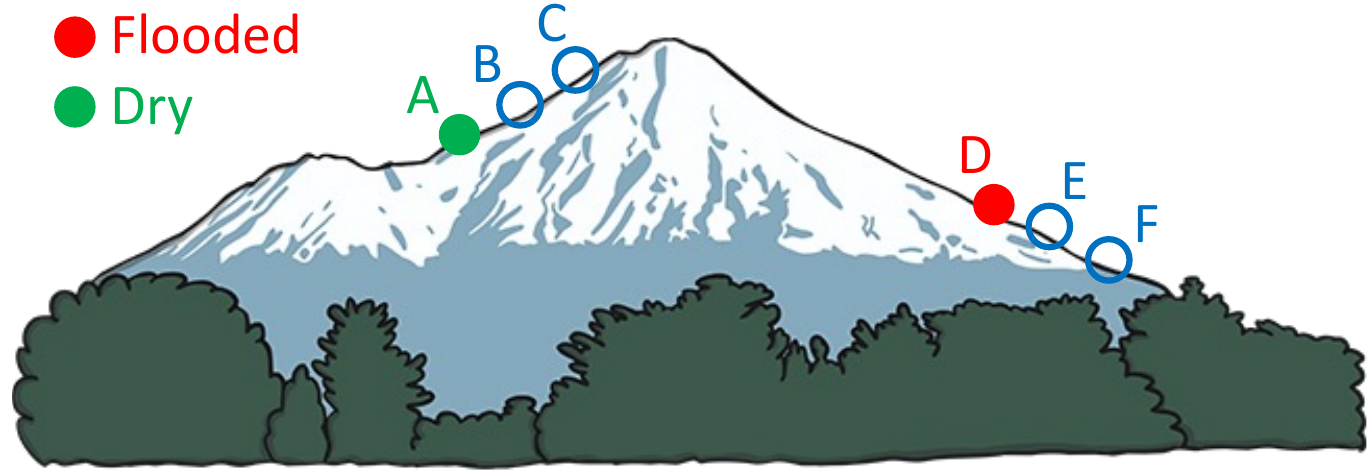} 
    \vspace{-2mm}
    \caption{Physical law of gravity: if pixel~$A$ is dry, then its higher adjacent pixel $B$ must be dry; while if pixel~$D$ is flooded, then its lower adjacent pixel $E$ must be flooded.}
    \label{mountain}
    \vspace{-4mm}
\end{figure}

Thanks to the availability of digital elevation model (DEM) data from sources such as USGS, existing deep learning approaches use the elevation map to improve segmentation performance by treating the map simply as an additional input channel of U-Net~\cite{unet} just like a regular spectral channel~\cite{DBLP:journals/staeors/BeniG21,konapala2021exploring,fraccaro2022deploying}. The physical law of gravity inherent to elevations as illustrated in Figure~\ref{mountain} (see its caption) is not encoded into the model design. 

Encoding physical law in model design can help better recover the true states (``flooded'' or ``dry'') of noisy pixels covered by clouds and tree canopies, which are often ambiguous just based on spectral features, and it has shown success in graphical models for flood mapping~\cite{DBLP:conf/kdd/0001S19,DBLP:journals/tkde/SainjuHJ22,DBLP:conf/aaai/SainjuH0020,hmt}. We propose to encode the physics-aware constraints into the more powerful deep learning models to achieve better 
performance.

Formally, given earth imagery with spectral features (e.g., RGB values) from optical satellites on a terrain surface (defined by an elevation map over a 2D grid), our goal is to conduct image segmentation so that each pixel (i.e., location) is classified as either ``flooded'' or ``dry''. 
In this paper, we propose an \underline{E}le\underline{va}tion-guided \underline{Net}work (EvaNet) for flood extent mapping on earth imagery, which is physics-guided (c.f.\ Figure~\ref{mountain}). Our main contributions are as follows:
\begin{itemize}
\item While a conventional loss function for image segmentation (e.g., pixel-wise cross-entropy loss $\mathcal{L}_{CE}$) is only computed from (i)~ground-truth pixel labels and (ii)~their predictions, we propose an elevation-guided regularization term $\mathcal{L}_{eva}$ to penalize adjacent pixel-pairs that violate the physical law of gravity. Interestingly, we find that $\mathcal{L}_{eva}$ alone can serve as a great loss function, not just as a regularization term.
\item We extensively explored various approaches to integrating the elevation map as an input to the convolution operations on the encoder and decoder sides of U-Net, and find that using an independent convolution branch over the elevation map (called the elevation path) to regulate the flow of spectral features by the gated linear unit (GLU)~\cite{glu} gives the best performance by striking a balance between model simplicity (to prevent overfitting) and model expressiveness.
\item Extensive experiments verify that our both techniques above are effective, and that when being combined, achieve the best performance beating U-Net and other non-deep-learning methods by a large margin. Moreover, we show that EvaNet is an effective drop-in replacement for U-Net to improve deep learning, flood mapping methods that originally rely on U-Net as a basic component.
\end{itemize}

\section{Related Work}\label{sec:related}
\noindent{\bf Image Segmentation.} Flood extent mapping is a semantic segmentation task, which classifies each pixel in an image into a semantic class (flood or dry in our problem). Besides semantic segmentation, there are other segmentation tasks such as instance segmentation~\cite{DBLP:conf/iccv/HeGDG17} which finds the object pixels in the bounding box of an object detected by an object detection model, and panoptic segmentation~\cite{DBLP:conf/cvpr/KirillovHGRD19} which combines semantic segmentation and instance segmentation. 

\vspace{2mm}
\noindent{\bf Flood Extent Mapping by Deep Learning.} A number of deep learning models for flood extent mapping have been explored by the remote sensing community recently. For example, 
\cite{DBLP:journals/staeors/BeniG21} first applies FCN~\cite{DBLP:conf/cvpr/LongSD15} 
on only the RGB data to obtain an initial flood map (Stage~1); then the work uses water level data from USGS combined with the DEM data to detect water pixels under tree canopy, by using a method called region growing (Stage~2); the result from Stage~2 is then used to improve the FCN-based flood extent for vegetated areas (Stage~3). We remark that elevation data was not used in Stage~1, and our EvaNet can serve as a drop-in replacement of FCN to improve the performance of Stage~1.

As another example, 
\cite{konapala2021exploring} uses spectral bands SWIR2, NIR and red values from Sentinel-2 to compute new indices as U-Net input, and DEM data is treated simply as an additional channel. Also, 
\cite{mateo2021towards} focuses on the application of onboard segmentation with small, nano satellites to reduce revisit time to disaster areas; they input 13-band Sentinel-2 image into U-Net for segmentation without using DEM data. While these works from the remote sensing community include new input modes that can extend our work, our EvaNet targets the more fundamental problem of effectively integrating DEM data with spectral features guided by the physical law of gravity, and can serve as a drop-in replacement of their na\"{i}ve U-Net or FCN models to improve performance.

\vspace{2mm}
\noindent{\bf Non-Deep-Learning Methods.} Hidden Markov tree (HMT)~\cite{hmt} is a state-of-the-art graphical model for flood extent mapping. HMT captures the directed spatial dependency based on flow directions across all locations by a reverse tree structure in the hidden class layer, which is similar to the hidden state sequence in hidden Markov models (HMM); spectral features are read out from the class state (dry or flooded) based on a learned multivariate Gaussian distribution. Later, hidden Markov contour tree (HMCT)~\cite{DBLP:conf/kdd/0001S19} further captures complex contour structures on a 3D surface. 

Both HMT and HMCT are for transductive learning where training samples are taken near the test region, or where a complete elevation map plus only limited spectral features in a target region are observed~\cite{DBLP:conf/aaai/SainjuH0020}.
More recently, inductive learning with HMT is explored with the help of U-Net. 
\cite{DBLP:journals/tkde/SainjuHJ22} proposes to use the inference stage of HMCT only as a post-processor of U-Net to cleanse its predicted noisy posterior class probability at all the pixels. The resulting model is called HMCT-PP. Our EvaNet can also serve as a drop-in replacement of the U-Net in HMCT-PP. 

Since the size of earth imagery on a terrain surface is often much larger than the input size of popular deep convolutional neural networks, segmentation is usually conducted on patches and the results are then stitched together. In contrast, HMT-based methods build a topological tree on the entire terrain surface so inference of the entire imagery is conducted together all at once. In HMCT-PP, the stitched U-Net predictions serve as the input to HMCT refinement.

\vspace{2mm}
\noindent{\bf Knowledge-Guided Machine Learning (KGML).} Unlike our problem that focuses on the Earth imagery at one snapshot to quickly map out flood-inundated area for disaster response, KGML research is usually formulated as a forecast problem fitting historical data or fitting a simulation model (e.g., based on PDE), so requires many temporal snapshots. The goal of KGML is more on geosciences to study climate and environmental phenomena/process for future disaster warning and mitigation, often using meteorological factors such as air temperature, precipitation, and wind speed. We discuss the related works on KGML in our online appendix~\cite{appendix}.

\section{Method}
In this section, we first formally define our problem and overview our techniques used in EvaNet. We then describe both techniques in detail.

\vspace{2mm}
\noindent{\bf Notations and Problem Definition.}\label{subsec:notation} 
We regard each image as a 2D grid $\mathbf{\Omega}$, and denote a pixel by $\mathbf{p}=(i, j)$, where $i$ and $j$ are the pixel coordinates in $\mathbf{\Omega}$. We also denote the elevation of pixel $\mathbf{p}$ by $h(\mathbf{p})$. 
Each pixel $\mathbf{p}$ has 8 adjacent pixels which constitute $\mathbf{p}$'s neighbor set, denoted by $\mathcal{N}(\mathbf{p})$. 
While pixels on the boundary of $\mathbf{\Omega}$ do not have 8 neighbor pixels, we pad the input image and its elevation map using the reflection of the input boundary, to ensure that all $\mathbf{p}\in\mathbf{\Omega}$ has 8 neighbors. 
We also use $\mathbf{p}_n$ to denote an arbitrary neighbor of $\mathbf{p}$, i.e., $\mathbf{p}_n\in\mathcal{N}(\mathbf{p})$. We call $(\mathbf{p}, \mathbf{p}_n)$ as a {\em pixel-pair}, where $\mathbf{p}$ and $\mathbf{p}_n$ are adjacent to each other in $\mathbf{\Omega}$. Finally, we define
 \begin{equation}\label{eq:delta_h}
\Delta h(\mathbf{p}, \mathbf{p}_n) = h(\mathbf{p}) - h(\mathbf{p}_n).
\end{equation}
In Figure~\ref{mountain}, we have $\Delta h(A, B)<0$ and $\Delta h(D, E)>0$. 

The goal of flood extent mapping is to assign each pixel $\mathbf{p}\in\mathbf{\Omega}$ a binary label indicating whether it is flooded or dry. 
Table~\ref{tab:notations} provides the list of notations used in this paper.

\begin{table}
\begin{center}
\begin{tabular}{ll}
    \hline
    {\bf Notation} & {\bf Meaning} \\
    \hline
    $\mathbf{\Omega}$ & The image as a 2D grid\\
    $\mathbf{p}$ & A pixel in $\mathbf{\Omega}$\\
    $\mathcal{N}(\mathbf{p})$ & The 8 neighboring pixels of $\mathbf{p}$ \\
    $\mathbf{p}_n$ & A pixel adjacent to $\mathbf{p}$\ \ ($\mathbf{p}_n\in\mathcal{N}(\mathbf{p})$)\\
    $h(\mathbf{p})$ & The elevation of $\mathbf{p}$ \\
    $\Delta h(\mathbf{p}, \mathbf{p}_n)$ & The elevation difference $h(\mathbf{p}) - h(\mathbf{p}_n)$\\
    $s_{flood}(\mathbf{p})$ & The predicted flood-score of $\mathbf{p}$ \\
    $s_{dry}(\mathbf{p})$ & The predicted dry-score of $\mathbf{p}$ \\
    \hline
\end{tabular}
\end{center}
\vspace{-2mm}
\caption{List of Notations}\label{tab:notations}
\end{table}
\setlength{\textfloatsep}{5pt}

\begin{figure}[!t]
    \centering
    \vspace{-5mm}
    \includegraphics[width=\columnwidth]{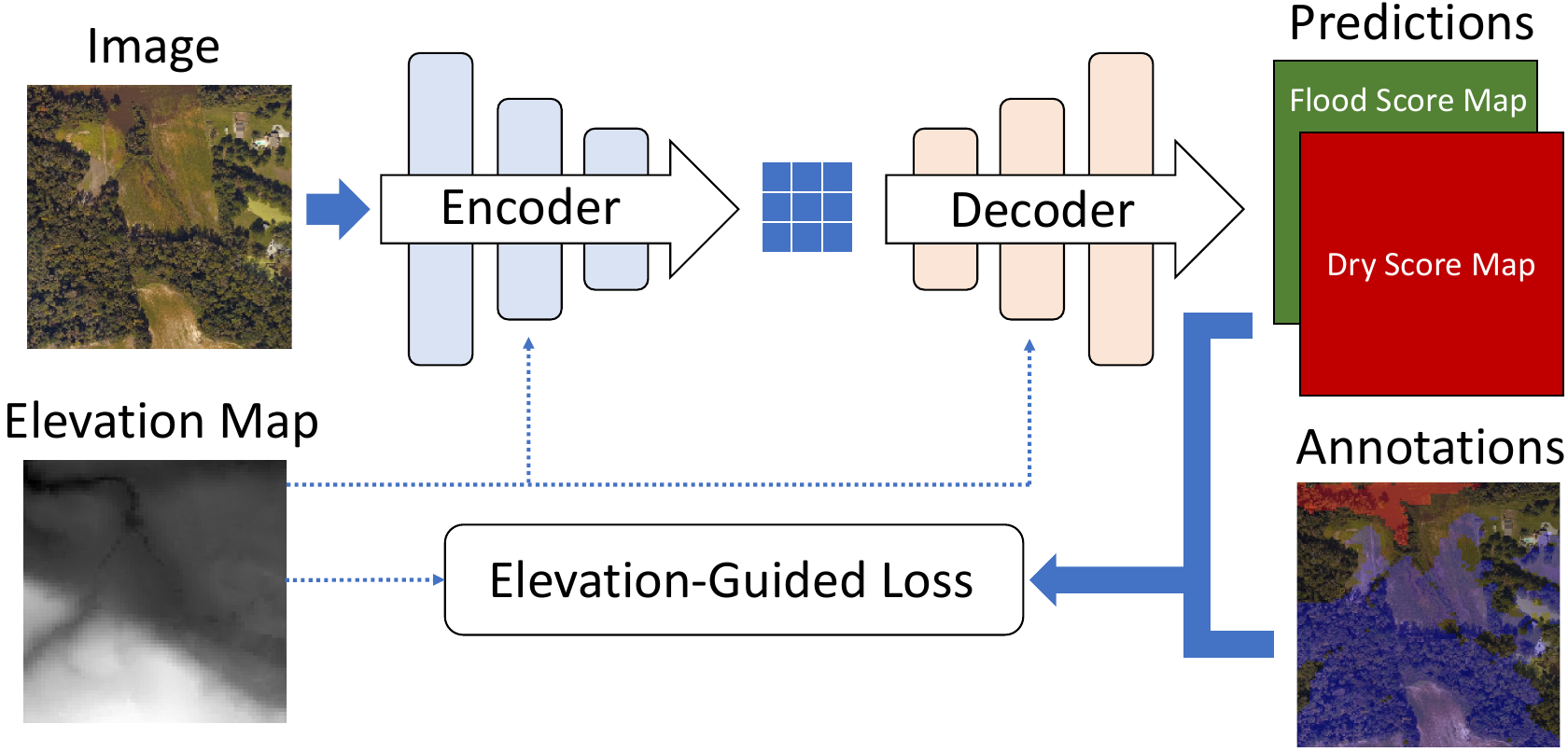}
	\caption{EvaNet architecture overview.}\label{overview}
\end{figure}

\vspace{2mm}
\noindent{\bf EvaNet Overview.} Figure~\ref{overview} overviews the architecture of EvaNet, where the encoder and decoder networks take not only the image but also the elevation map in convolution computation, which we will explain in section ``Elevation-Regulated Convolutions"; and where the loss function takes not only the predictions and the ground-truth annotation, but also the elevation map in the loss computation, which we will explain in section~``Elevation-Guided Loss".

\subsection{Elevation-Guided Loss}\label{ssec:loss}
Given an input image of size $W\times H$, a flood segmentation model (U-Net or EvaNet) outputs a tensor of size $W\times H\times 2$ with 2 channels: one is a dry score map $s_{dry}\in\mathbb{R}^{W\times H}$, and the other is a flood score map $s_{flood}\in\mathbb{R}^{W\times H}$, as illustrated in Figure~\ref{overview}. To compute the loss function, we apply pixel-wise softmax where the probability that a pixel $\mathbf{p}$ is flooded (resp.\ dry) is given by $p_{flood}(\mathbf{p})=\frac{e^{s_{flood}(\mathbf{p})}}{e^{s_{flood}(\mathbf{p})} + e^{s_{dry}(\mathbf{p})}}$ \big(resp.\ $p_{dry}(\mathbf{p})=\frac{e^{s_{dry}(\mathbf{p})}}{e^{s_{flood}(\mathbf{p})} + e^{s_{dry}(\mathbf{p})}}$\big)\footnote{Alternatively, since our label is binary, one may let the segmentation model output only flood score map $s_{flood}$, and compute $p_{flood}=\text{sigmoid}(s_{flood})$ and $p_{dry}=1-p_{flood}$. However, we find this formulation to be less numerically stable in PyTorch than using softmax.}.

We define the ground-truth label of a pixel $\mathbf{p}$ as follows:
{\small
$$gt(\mathbf{p})=\left\{
\begin{aligned}
1& \ \ \ \ \ \ \ \ & \text{$\mathbf{p}$ is flooded,\ \ }\\
0& \ \ \ \ \ \ \ \ & \text{$\mathbf{p}$ is unlabeled,}\\
-1& \ \ \ \ \ \ \ \ & \text{$\mathbf{p}$ is dry.\ \ \ \ \ \ }
\end{aligned}
\right.
$$
}
\!\!Note that a pixel may be unlabeled since it is covered by tree canopy. 
We provide domain experts with a 3D visualization tool where they can navigate the terrain surface by shifting and rotation, and where they can annotate the pixels using tools like `brush' if they are confident of the pixel labels. Elevation-guided breadth-first search (BFS)~\cite{saugat_demo} was adopted to speed up annotation, where (i)~when an annotator marks an individual pixel $\mathbf{p}$ as flooded, the label propagates to nearby pixels with lower elevations by `pit-filling' breadth-first search stopping when reaching pixels with elevation higher than $h(\mathbf{p})$, and (ii)~when an annotator marks an individual pixel $\mathbf{p}$ as dry, the label propagates to nearby pixels by `hill-climbing' breadth-first search stopping when reaching pixels with elevation starting to drop. 
The labels of some pixels covered by tree canopy may be derived in this way from nearby pixels using the physical law of gravity.

U-Net uses only pixel-wise cross-entropy loss:
$$\mathcal{L}_{CE}=-\!\!\!\!\!\!\!\!\sum_{\mathbf{p}\in\mathbf{\Omega}\ \wedge\ gt(\mathbf{p})=1}\!\!\!\!\!\!\!\!\log p_{flood}(\mathbf{p})\ -\!\!\!\!\!\!\!\!\sum_{\mathbf{p}\in\mathbf{\Omega}\ \wedge\ gt(\mathbf{p})=-1}\!\!\!\!\!\!\!\!\log p_{dry}(\mathbf{p}).$$

\vspace{2mm}
\noindent{\bf Elevation-Guided Regularization.} However, $\mathcal{L}_{CE}$ does not take the pixel elevations into consideration. To utilize the elevation map, we propose a regularization term $\mathcal{L}_{eva}$ to penalize predictions that violate the physical constraints shown in Figure~\ref{mountain}. Unlike $\mathcal{L}_{CE}$ which is computed over individual pixels $\mathbf{p}$, $\mathcal{L}_{eva}$ is computed over individual pixel-pairs $(\mathbf{p}, \mathbf{p}_n)$:
\begin{equation}\label{eq:eva}
\mathcal{L}_{eva}\ =\!\!\!\!\!\!\!\!\sum_{\mathbf{p}\in\mathbf{\Omega}\ \wedge\ gt(\mathbf{p})\neq0} \ \ \sum_{\mathbf{p}_n\in\mathcal{N}(\mathbf{p})}\Big(w(\mathbf{p}, \mathbf{p}_n) \cdot \delta(\mathbf{p}, \mathbf{p}_n)\Big),
\end{equation}
where (i)~$w(\mathbf{p}, \mathbf{p}_n)$ is a weight term indicating how much the pixel-pair $(\mathbf{p}, \mathbf{p}_n)$ is relevant to the physical constraints, and (ii)~$\delta(\mathbf{p}, \mathbf{p}_n)$ measures how much the label predictions of $\mathbf{p}$ and $\mathbf{p}_n$ deviate from the physical constraints. 
We next explain the formulations of $w(\mathbf{p}, \mathbf{p}_n)$ and $\delta(\mathbf{p}, \mathbf{p}_n)$.

Note that if a pixel $\mathbf{p}$ is unlabeled (i.e., $gt(\mathbf{p})=0$), it has no term in the summation formulations of $\mathcal{L}_{CE}$ and $\mathcal{L}_{eva}$, so it will not provide any supervision during training. 

\vspace{2mm}
\noindent{\bf Formulation of $\delta(\mathbf{p}, \mathbf{p}_n)$.} We define $\delta(\mathbf{p}, \mathbf{p}_n)$ as follows, and we will explain its intuition soon:
\begin{equation}\label{eq:delta}
\delta(\mathbf{p}, \mathbf{p}_n) = 1-gt(\mathbf{p}_n)\cdot f(\mathbf{p}),
\end{equation}
where $f(\mathbf{p})\in(-1, 1)$ is defined as follows:
$$f(\mathbf{p})=\left\{
\begin{aligned}
\text{sigmoid}(s_{flood}(\mathbf{p}))& \ \ \ & f_{flood}(\mathbf{p}) \geq f_{dry}(\mathbf{p}),\ \ \\
-\,\text{sigmoid}(s_{dry}(\mathbf{p}))& \ \ \ & f_{flood}(\mathbf{p}) < f_{dry}(\mathbf{p}).\ \ 
\end{aligned}
\right.
$$
In other words, the higher confidence that our prediction is for $\mathbf{p}$ to be flooded (resp.\ dry), the closer $f(\mathbf{p})$ is to 1 (resp.\ $-1$). An alternative definition that achieves this goal is to let the segmentation model output only a flood score map $s_{flood}$, and compute $f(\mathbf{p})=\tanh(s_{flood})$. However, we find this formulation less numerically stable in PyTorch.

\vspace{2mm}
\noindent{\bf Formulation of $w(\mathbf{p}, \mathbf{p}_n)$.} We define $w(\mathbf{p}, \mathbf{p}_n)$ as follows, and we will explain its intuition soon:
\begin{equation}\label{eq:w1}
w(\mathbf{p}, \mathbf{p}_n) \!\,=\!\, \mathbf{1}\{-gt(\mathbf{p}_n) \!\,\cdot\!\, \Delta h(\mathbf{p}, \mathbf{p}_n)>0\},
\end{equation}
where $\mathbf{1}\{.\}$ is an indicator variable that equals 1 if the condition holds, and 0 otherwise. 
Note that only $\delta(\mathbf{p}, \mathbf{p}_n)$ involves prediction since it uses $f(\mathbf{p})$; in contrast, $w(\mathbf{p}, \mathbf{p}_n)$ is purely decided by $gt(\mathbf{p}_n)$, $h(\mathbf{p})$ and $h(\mathbf{p}_n)$, i.e., annotations and elevations. 
Also note that if $\mathbf{p}_n$ is unlabeled (i.e., $gt(\mathbf{p}_n)=0$), then $w(\mathbf{p}, \mathbf{p}_n)=0$ so the term in the summation formulation of $\mathcal{L}_{eva}$ in Eq~(\ref{eq:eva}) is 0, i.e., pixel-pair $(\mathbf{p}, \mathbf{p}_n)$ provides no supervision during training.

\begin{figure}[t]
    \centering
    \includegraphics[width=\columnwidth]{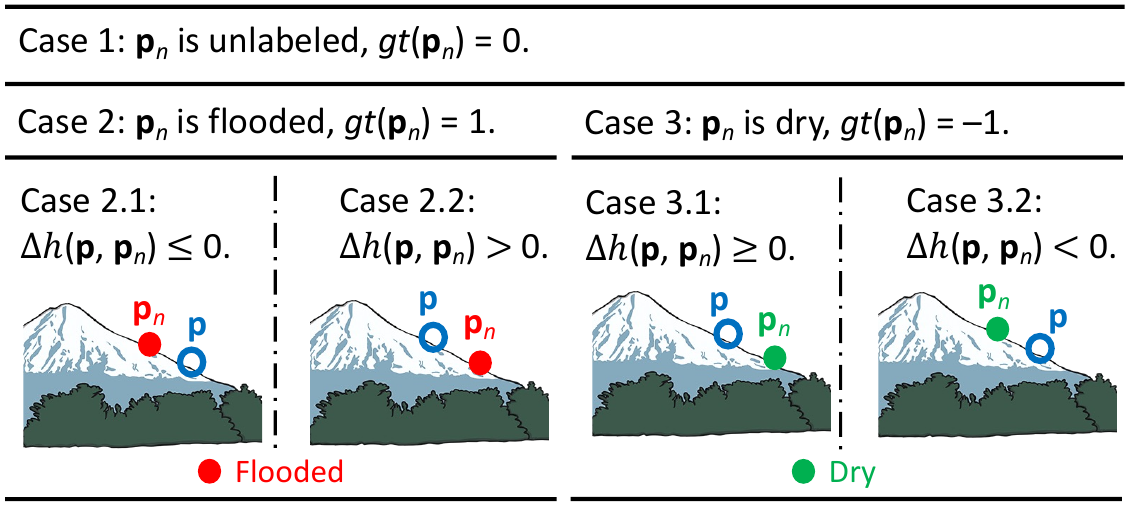}
    \caption{Five cases for pixel-pair $(\mathbf{p}, \mathbf{p}_n)$.}\label{cases}
\end{figure}

\vspace{2mm}
\noindent{\bf Intuition Behind the Formulations.} We next provide the intuition of our formulations for $\delta(\mathbf{p}, \mathbf{p}_n)$ and $w(\mathbf{p}, \mathbf{p}_n)$. We divide the possible scenarios for a pixel-pair $(\mathbf{p}, \mathbf{p}_n)$ into 5 cases shown in Figure~\ref{cases}, based on $gt(\mathbf{p}_n)$ as the primary partitioning key and $\Delta h(\mathbf{p}, \mathbf{p}_n)$ as the secondary key.

The physical constraints illustrated in Figure~\ref{mountain} correspond to 2 sub-cases in Figure~\ref{cases}: (1)~Case~2.1 where we can derive that $\mathbf{p}$ should also be flooded, and (2)~Case~3.1 where $\mathbf{p}$ should also be dry. Note that based on Eq~(\ref{eq:w1}), $w(\mathbf{p}, \mathbf{p}_n)=1$ for these 2 cases, and $w(\mathbf{p}, \mathbf{p}_n)=0$ for the other 3 cases where $\mathbf{p}$ can be either dry or flooded without violating any constraints. As a result, $w(\mathbf{p}, \mathbf{p}_n)$ in Eq~(\ref{eq:w1}) acts as a switch that is on only for those pixel-pairs $(\mathbf{p}, \mathbf{p}_n)$ that belong to Case~2.1 and Case~3.1, and excludes the other pixel-pairs when computing $\mathcal{L}_{eva}$ using Eq~(\ref{eq:eva}). We call this weighting scheme given by Eq~(\ref{eq:w1}) as {\bf binary weighting}.

Based on Eq~(\ref{eq:delta}), (1)~$\delta(\mathbf{p}, \mathbf{p}_n)=1-f(\mathbf{p})$ in Case~2.1, so if $f(\mathbf{p})$ is close to 1 (i.e., predicted closer to be flooded) as aligned with the physical constraint, then $\delta(\mathbf{p}, \mathbf{p}_n)$ is close to 0; while if $f(\mathbf{p})$ is close to $-1$ (i.e., predicted closer to be dry) which violates the physical constraint, then $\delta(\mathbf{p}, \mathbf{p}_n)$ is close to the supremum 2 so the penalty term in $\mathcal{L}_{eva}$ is maximized. Also, (2)~$\delta(\mathbf{p}, \mathbf{p}_n)=1+f(\mathbf{p})$ in Case~3.1 by Eq~(\ref{eq:delta}), so if $f(\mathbf{p})$ is close to $-1$ (i.e., predicted closer to be dry) as aligned with the physical constraint, then $\delta(\mathbf{p}, \mathbf{p}_n)$ is close to 0; while if $f(\mathbf{p})$ is close to 1 (i.e., predicted closer to be flooded) which violates the physical constraint, then $\delta(\mathbf{p}, \mathbf{p}_n)$ is close to 2 to maximize the penalty term.

Figure~\ref{Bar_1_3} in our online appendix~\cite{appendix} shows the pixel distribution among the 5 cases on our datasets, where Cases~2.1 and~3.1 are dominating where the physical constraints apply, so we expect sufficient regularization.

\begin{figure*}[t]
	\centering
	\includegraphics[width=1.8\columnwidth]{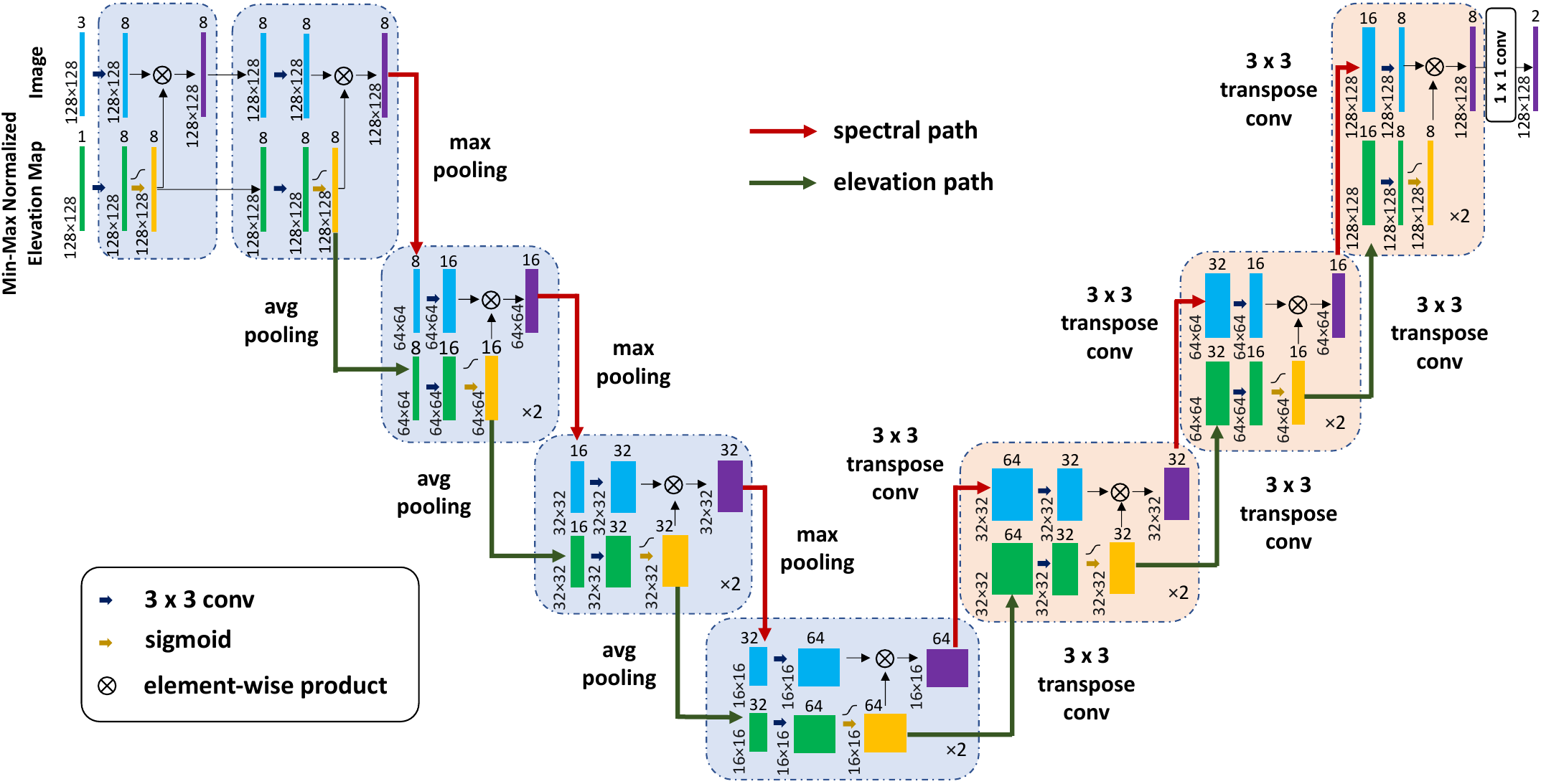}
	\vspace{-2mm}
	\caption{EvaNet encoder-decoder network architecture.}\label{arch}
\vspace{-4mm}
\end{figure*}

\vspace{2mm}
\noindent{\bf Alternative Formulations of $w(\mathbf{p}, \mathbf{p}_n)$.} The binary weighting scheme in Eq~(\ref{eq:w1}) only switches on the pixel-pairs satisfying Case~2.1 and Case~3.1, and treats their penalty terms as equally important. We also explored the adjustment of weights based on $|\Delta h(\mathbf{p}, \mathbf{p}_n)|$. The intuition is that, for example in Case~2.1, the higher $\mathbf{p}_n$ is than $\mathbf{p}$, the more penalty weight will be imposed if $\mathbf{p}$ is misclassified as dry.

Based on this idea, we define the {\bf elevation-difference weighting scheme} as follows:
\begin{equation}\label{eq:w2}
w(\mathbf{p}, \mathbf{p}_n) \!\,=\!\, \max\{-gt(\mathbf{p}_n) \!\,\cdot\!\, \Delta h(\mathbf{p}, \mathbf{p}_n), 0\},
\end{equation}
so that $w(\mathbf{p}, \mathbf{p}_n)=|\Delta h(\mathbf{p}, \mathbf{p}_n)|$ for Cases~2.1 and~3.1, and 0 otherwise. We also explored a {\bf log-elevation-difference weighting scheme} where 
$w(\mathbf{p}, \mathbf{p}_n)=\log\big(1+|\Delta h(\mathbf{p}, \mathbf{p}_n)|\big)$ for Cases~2.1 and~3.1, and 0 otherwise:
\begin{equation}\label{eq:w3}
w(\mathbf{p}, \mathbf{p}_n) \!\,=\!\, \log\!\Big(1 \!\,+\!\, \max\{-gt(\mathbf{p}_n) \!\,\cdot\!\, \Delta h(\mathbf{p}, \mathbf{p}_n), 0\}\Big).\ \ \!\!\!\!
\end{equation}

However, we find that these weighting schemes are not better than binary weighting. The results of comparison can be found in Table \ref{table:weighting_scheme_comparison} of the appendix~\cite{appendix}. 

\vspace{2mm}
\noindent{\bf Implementing $\mathcal{L}_{eva}$.} Recall that Eq~(\ref{eq:eva}) sums over pixel-pairs rather than individual pixels, but both the predicted score maps $s_{flood}$ and $s_{dry}$ and the ground-truth map $gt$ are on pixels. Fortunately, PyTorch supports an ``unfold'' operation that can expand each pixel to include its local $3\times3$ pixel-neighborhood. We utilize ``unfold'' to expand each pixel $\mathbf{p}$ in $s_{flood}$, $s_{dry}$ and $gt$ with its neighbors $\mathbf{p}_n$ to facilitate the computation of $w(\mathbf{p}, \mathbf{p}_n)$ and $\Delta h(\mathbf{p}, \mathbf{p}_n)$.

\vspace{2mm}
\noindent{\bf Loss Schemes.} Since $\mathcal{L}_{eva}$ is formulated as a regularization term that penalizes pixel-pairs $(\mathbf{p}, \mathbf{p}_n)$ in Cases~2.1 and~3.1 based on how much the label prediction of $\mathbf{p}$ violates the physical constraints, a natural loss function is given by
\begin{equation}\label{eq:loss_form}
\mathcal{L}=\mathcal{L}_{CE} + \lambda\cdot\mathcal{L}_{eva},
\end{equation}
where hyperparameter $\lambda$ balances the importance between $\mathcal{L}_{CE}$ and $\mathcal{L}_{eva}$. 
U-Net simply uses $\mathcal{L}=\mathcal{L}_{CE}$ (i.e., $\lambda=0$). By setting $\lambda=1$, we obtain a representative hybrid loss $\mathcal{L}=\mathcal{L}_{CE} + \mathcal{L}_{eva}$. Another extreme is using $\mathcal{L}=\mathcal{L}_{eva}$ (i.e., $\lambda=\infty$), which could be a bit surprising since we exclude $\mathcal{L}_{CE}$ and only use the regularization term to penalize the violation of physical constraints. Surprisingly, this last loss turns out to perform well in our experiments (c.f., Table~\ref{table:loss_comparison} in our  appendix~\cite{appendix}), probably because Cases~2.1 and~3.1 are well spread out on the entire terrain (c.f., Figure~\ref{Bar_1_3} in our appendix~\cite{appendix}). In other words, $\mathcal{L}_{eva}$ by itself serves as a good loss function for flood mapping.

\subsection{Elevation-Regulated Convolutions}\label{ssec:conv}
Figure~\ref{arch} shows the architecture of EvaNet's encoder-decoder network, where the encoder and decoder each consists of 3 convolution blocks, each with two novel elevation-regulated convolution (ERC) layers to be introduced soon.

EvaNet takes two inputs: (1)~a $128\times128$ image patch and (2)~the associated elevation map of the patch. Here, we partition a large high-resolution satellite image into $128\times128$ patches, since an entire satellite image is too large to input into any convolution neural network given the current GPU memory capacity. For this purpose, we pad each image on its boundary using the reflection mode, so that both the width and height of the padded image is divisible by 128. For example, for an $1856\times4104$ input image, we pad 32 pixels on the left and right boundaries, and pad 60 pixels on the upper and lower boundaries, giving a padded image of size $1920\times 4224$ which gives $15\times33$ patches.

Since neural networks take normalized inputs, we conduct min-max normalization of the elevations and input the normalized elevation patch to our encoder. Note that in contrast, raw elevations are used when we calculate $w(\mathbf{p}, \mathbf{p}_n)$ (using Eq~(\ref{eq:w1}), Eq~(\ref{eq:w2}) or Eq~(\ref{eq:w3})) to compute $\mathcal{L}_{eva}$.

As Figure~\ref{arch} shows, each convolution block consists of two ERC layers. Each ERC layer takes a spectral tensor (blue) $\mathbf{X}$ and an elevation-derived tensor (green) $\mathbf{X}_e$, and outputs a regulated spectral tensor (purple) $\mathbf{Y}$ and a gated elevation-derived tensor (yellow) $\mathbf{Y}_e$. Specifically, both inputs first pass through a $3\times3$ convolution, and the two resulting tensors are combined by gated linear unit (GLU)~\cite{glu}:
$\mathbf{Y}_e=\text{sigmoid}\big(\text{conv2d}(\mathbf{X}_e)\big)$, $\mathbf{Y}=\text{conv2d}(\mathbf{X})\otimes\mathbf{Y}_e,$
where $\text{conv2d}(.)$ is a $3\times3$ convolution with a `replicate' padding mode, and $\otimes$ is element-wise product. The outputs ($\mathbf{Y}$, $\mathbf{Y}_e$) then serve as the inputs to the next ERC layer. 

Intuitively, $\mathbf{Y}_e\in(0, 1)^{H\times W\times C}$ serves as an elevation-based gate to control the flow of $\text{conv2d}(\mathbf{X})$ to $\mathbf{Y}$, and since $\mathbf{Y}_e$ is computed from $\mathbf{X}_e$ with a $3\times 3$ convolution, it is able to capture the local terrain topology. We choose GLU for elevation-based regulation since it is found to outperform other gating mechanisms in recent works~\cite{DBLP:conf/ijcai/WuPLJZ19,DBLP:conf/kdd/Hui0CK21,DBLP:conf/icdm/Hui0CK21}. We also tested other alternative gating mechanisms such as~\cite{arevalo2017gated,kim2018robust} and found GLU to perform the best. The related experiments are reported in Table~\ref{table:fusiopn_comparison_4c_7c} of our online appendix~\cite{appendix}.

Also, we use `replicate' padding for $\text{conv2d}(.)$ to ensure the quality of predictions on patch boundary. This is in contrast to zero padding, which we tested and found that the abrupt value drops at the patch boundary lead to poor prediction quality for pixels at the boundary.

As Figure~\ref{arch} shows, at the encoder side, $\mathbf{Y}$ is downsampled using $2\times 2$ max pooling which is commonly used for spectral tensors, and $\mathbf{Y}_e$ is downsampled using $2\times 2$ average pooling as a lower-resolution terrain approximation. We tried other pooling combinations but this setting gives the best predictions as demonstrated in Table \ref{table:pooling_scheme_comparison} of the online appendix~\cite{appendix}. At the decoder side, we use $3\times3$ transpose convolution for both $\mathbf{Y}$ and $\mathbf{Y}_e$.

\vspace{-2mm}
\section{Experiments \& Results}\label{sec:results}
\begin{table*}[t]
 \centering
 \fontsize{9}{11}\selectfont
 \begin{tabular}{c c c c c c c}
 \hline
    \multicolumn{1}{c||}{} & Height & Width & Region & \%Dry & \%Flood & \%Annotated\\ 
    \midrule
    \multicolumn{1}{c||}{R1}  &  1856 & 4104 & Grimesland, NC & 19.97\% & 41.22\% & 61.19\%\\
    \multicolumn{1}{c||}{R2}  &  2240 & 4704 & Greenville-Central, NC & 52.32\% & 34.43\% & 86.75\%\\
    \multicolumn{1}{c||}{R3}  &  3136 & 6472 & Falkland, NC & 58.35\% & 26.54\% & 84.89\%\\ 
    \multicolumn{1}{c||}{R4}  &  2782 & 5500 & Kinston, NC & 49.20\% & 40.95\% & 90.15\%\\ 
    \multicolumn{1}{c||}{R5}  &  2800 & 5400 & Greenville-West, NC & 45.18\% & 27.63\% & 72.81\%\\
    \multicolumn{1}{c||}{R6}  &  2828 & 5286 & Baytown, TX & 37.56\% & 39.48\% & 77.04\%\\
    \multicolumn{1}{c||}{R7}  &  2306 & 4802 & Richmond, TX & 44.44\% & 36.44\% & 80.88\%\\
    \bottomrule
    \end{tabular}
    \caption{Regions and their statistics.}
    \label{tab:data}
    \vspace{-4mm}
\end{table*}
%

%
\noindent{\bf Data.} We obtain high-resolution aerial imagery from NOAA National Geodetic Survey during Hurricane Matthew in North Carolina (NC) in 2016~\cite{Hurrican82:online}. The accompanied DEM data are obtained from the University of North Carolina Libraries~\cite{ncsu}. Futhermore, to test if our models trained on the Matthew 2016 dataset can generalize to a different flooding event, we also obtained  two more regions in Baytown, TX, and Richmond, TX during Hurricane Harvey in 2017 from~\cite{Hurrican61:online}, and the corresponding DEM data was obtained from USGS data downloader~\cite{TNMDownl88:online}. The aerial imagery from both events are 0.3~m $\times$ 0.3~m in resolution, the DEM data for Matthew 2016 are 6~m $\times$ 6~m and for Harvey 2017 are 3~m $\times$ 3~m in resolution.  We resampled them into a resolution of 2~m $\times$ 2~m, which is fine enough for the purpose of flood extent mapping.
Table~\ref{tab:data} shows the regions we used, where R1--R5 are from Matthew 2016, and R6--R7 is from Harvey 2017. While high-resolution Earth imagery and DEM data are abundant, the ground-truth flood maps need to be annotated by domain experts. So, we provide a 3D visualization tool to domain experts for annotating pixels. 


In our online appendix~\cite{appendix}, Figure~\ref{kinston} shows the imagery for Kinston, NC during Hurricane Matthew in 2016, its annotation where pixels are labeled as flooded or dry, and the normal-time imagery from Google Earth. We expect that inputting the normal-time image into a model is effective in helping identify flooded regions by providing a contrast. This is confirmed our experiments reported in Table~\ref{table:model_comparison_4c_7c} of our online appendix, which also aligns with findings from prior works~\cite{drakonakis2022ombrianet,rudner2019multi3net}. 
Some pixels are ambiguous (e.g., covered with tree canopies) so remain unlabeled. Unlabeled pixels are not included in the loss computation during training, and not included during test when calculating the evaluation measures. As Table~\ref{tab:data} shows, most pixels of each region are annotated. Without loss of generalization, in Table~\ref{tab:data}, we use R1 and R2 for training, and use R3, R4, R5, R6 and R7 for test.

\begin{table*}[t]
 \centering
 \fontsize{9}{11}\selectfont
 \begin{tabular}{c | c|| c c c c | c c c c }
    \toprule
    \multicolumn{1}{c|}{\multirow{2}{*}{Region}} & \multicolumn{1}{c||}{\multirow{2}{*}{Method}}& \multicolumn{4}{c|}{Dry} & \multicolumn{4}{c}{Flood} \\
    \cline{3-10}
     & & Accuracy & Precision & Recall & F1-Score & Accuracy & Precision & Recall & F1-Score \\ 
     \midrule
     \multicolumn{1}{c|}{\multirow{4}{*}{R3: Falkland, NC}} & \multicolumn{1}{c||}{HMT} & 91.91 & 95.31 & 92.80 & 94.03 & 91.91 & 85.03 & 89.95 & 87.42 \\ 
     & \multicolumn{1}{c||}{U-Net 3C}  & 81.26 & 95.48 & 76.34 & 84.85 & 81.26 & 63.90 & 92.05 & 75.44 \\ 
     & \multicolumn{1}{c||}{U-Net 7C}  & 94.94  & 95.91 & 96.77 & 96.34 & 94.94 & 92.76 & 90.93 & 91.83 \\
     & \multicolumn{1}{c||}{EvaNet 7C}  &  \textbf{97.15} &  \textbf{97.63} & \textbf{98.23} & \textbf{97.93} &  \textbf{97.15} & \textbf{96.06}  & \textbf{94.77} & \textbf{95.41}\\ 
     \midrule
     \multicolumn{1}{c|}{\multirow{4}{*}{R4: Kinston, NC}} & \multicolumn{1}{c||}{HMT}  &  84.33 & 81.14 & 92.88 & 86.61 & 84.33 & 89.64 & 74.06 & 81.11\\ 
     & \multicolumn{1}{c||}{U-Net 3C}  &  90.70 & \textbf{93.57} & 89.08 & 91.27 & 90.70 & 87.60 & \textbf{92.64} & 90.05\\ 
     & \multicolumn{1}{c||}{U-Net 7C}  &  83.90 & 78.45 & 97.20 & 86.82 & 83.90  & 95.28 & 67.92 & 79.31 \\  
     & \multicolumn{1}{c||}{EvaNet 7C}  & \textbf{92.16} & 88.64 & \textbf{98.22} & \textbf{93.18} & \textbf{92.16} & \textbf{97.54} & 84.88 & \textbf{90.77} \\ 
     \midrule
     \multicolumn{1}{c|}{\multirow{4}{*}{R5: Greenville-West, NC}} & \multicolumn{1}{c||}{HMT}  &  61.91 & 62.21 & \textbf{98.39} & 76.23 & 61.91 & 46.18 & 2.25 & 4.30 \\ 
     & \multicolumn{1}{c||}{U-Net 3C}  &  86.12 & 96.93 & 80.18 & 87.76 & 86.12 & 74.73 & 95.84 & 83.98 \\ 
     & \multicolumn{1}{c||}{U-Net 7C}  & 95.97 & 98.49 & 94.96 & 96.70 & 95.97 & 92.21 & 97.62 & 94.84\\ 
     & \multicolumn{1}{c||}{EvaNet 7C}  & \textbf{97.75} & \textbf{98.65} & 97.70 & \textbf{98.18} & \textbf{97.75} & \textbf{96.30} & \textbf{97.81} & \textbf{97.05}\\ 
     \midrule
     \multicolumn{1}{c|}{\multirow{4}{*}{R6: Baytown, TX}} & \multicolumn{1}{c||}{HMT}  &  63.30 & 57.08 & \textbf{99.65} & 72.59 & 63.30 & \textbf{98.86} & 28.73 & 44.52\\ 
     & \multicolumn{1}{c||}{U-Net 3C}  &  63.68 & 88.50 & 29.31 & 44.0 & 63.68 & 58.90 & 96.38 & 73.12 \\ 
     & \multicolumn{1}{c||}{U-Net 7C}  & 74.41 & 88.76 & 54.40 & 67.46 & 74.41 & 68.30 & 93.45 & 78.90 \\ 
     & \multicolumn{1}{c||}{EvaNet 7C}  & \textbf{91.50} & \textbf{98.46} & 83.89 & \textbf{90.59} & \textbf{91.50} & 86.56 & \textbf{98.75} & \textbf{92.26} \\ 
     \midrule
     \multicolumn{1}{c|}{\multirow{4}{*}{R7: Richmond, TX}} & \multicolumn{1}{c||}{HMT}  &  74.67 & 68.61 & \textbf{99.38} & 81.17 & 76.67 & \textbf{98.33} & 44.55 & 61.32\\ 
     & \multicolumn{1}{c||}{U-Net 3C}  &  53.95 & 81.36 & 20.99 & 33.38 & 53.95 & 49.42 & \textbf{94.13} & 64.81 \\ 
     & \multicolumn{1}{c||}{U-Net 7C}  & 56.13 & 74.17 & 30.92 & 43.65 & 56.13 & 50.77 & 86.87 & 64.08 \\ 
     & \multicolumn{1}{c||}{EvaNet 7C}  & \textbf{84.06} & \textbf{90.07} & 79.79 & \textbf{84.62} & \textbf{84.06} & 78.37 & 89.28 & \textbf{83.46} \\ 
     \bottomrule
     \end{tabular}
   \vspace{-1mm}
   \caption{Model comparison (unit: \%).}
   \label{table:model_comparison}
   \vspace{-1mm}
\end{table*}

\begin{table*}[t]
 \centering
 \fontsize{9}{11}\selectfont
 \begin{tabular}{c || c c c c|c c c c}
    \toprule
    \multirow{2}{*}{Method} & \multicolumn{4}{c|}{Dry} & \multicolumn{4}{c}{Flood}\\ 
     \cline{2-9}
     & Accuracy & Precision & Recall & F1-Score  & Accuracy & Precision & Recall & F1-Score  \\ 
     \midrule
     \multicolumn{1}{c||}{U-Net 7C}  & 83.90 & 78.45 & 97.20 & 86.82 & 83.90 & 95.28 & 67.92 & 79.31  \\ 
     \multicolumn{1}{c||}{U-Net  7C w/ Eva-Conv}  & 85.56 & 79.75 & 98.56 & 88.17 & 85.56 & 97.61 & 69.93 & 81.48  \\  
     \multicolumn{1}{c||}{U-Net  7C w/ Eva-Reg}  & 86.59 & 80.96 & \textbf{98.64} & 88.93 & 86.59 & \textbf{97.78} & 72.13 & 83.02   \\ 
     \multicolumn{1}{c||}{EvaNet  7C}  & \textbf{92.16} & \textbf{88.64} & 98.22 & \textbf{93.18} & \textbf{92.16} & 97.54 & \textbf{84.88} & \textbf{90.77}  \\ 
     \bottomrule
      \end{tabular}
   \vspace{-1mm}
   \caption{Ablation study (unit: \%) on test region R4.}
   \label{table:ablation_study}
   \vspace{-4mm}
\end{table*}

\vspace{2mm}
\noindent{\bf Models.} We consider three U-Net baselines: (1)~``U-Net 3C'' with input using the RGB channels of the disaster-time image; (2)~``U-Net 4C'' with input adding an additional channel for the elevation map; (3)~``U-Net 7C'' with input also adding the RGB channels of the normal-time image. 

For EvaNet, we consider two versions: (1)~``EvaNet 4C'' using only the disaster-time RGB image for the leftmost cyan `image' input tensor shown in Figure~\ref{arch}, and (2)~``EvaNet 7C'' using both disaster-time and normal-time images for that input tensor. We remark that our new elevation-regulated convolution (denoted by {\bf Eva-Conv}) and elevation-guided loss (denoted by {\bf Eva-Reg}) can serve as drop-in replacements of the regular convolution and binary cross-entropy loss in any encoder-decoder model for semantic segmentation, and we choose the popular U-Net model in this work just as a representative. 

For physics-guided graphical models, we compare with hidden Markov tree (HMT)~\cite{hmt,DBLP:conf/kdd/0001S19}. Since HMT uses transductive learning, we use the same region for training and test. For U-Net and EvaNet, we use two regions R1 and R2 in Table~\ref{tab:data} for training and the other regions for test. For example, when reporting the performance measures on R3, HMT is trained on R3 and tested on R3, while all other models are trained with the patches of R1 and R2, and then tested on patches of R3 and stitched.

By default, EvaNet uses only $\mathcal{L}_{eva}$ in Eq~(\ref{eq:eva}) as the loss function, where $w(\mathbf{p}, \mathbf{p}_n)$ uses the binary weighting scheme in Eq~(\ref{eq:w1}). We use 3 convolution blocks of Eva-Conv for both the encoder and the decoder as shown in Figure~\ref{arch}, where the input has 7 channels (disaster-time and normal-time RGB plus elevation map), and the encoder downsamples the spectral path using max pooling, and downsamples the elevation path using average pooling. 
These default configurations are found to be the best-performing, as demonstrated by our experiments in the online appendix~\cite{appendix} (see Tables~\ref{table:loss_comparison}--\ref{table:model_comparison_4c_7c}). 
We trained our EvaNet models and the U-Net baselines on a cluster with NVIDIA-P100 GPUs for 100 epochs with a learning rate of $1\times10^{-7}$ and batch size of 4. 
These default settings are found to be the best-performing through  extensive tests.

\vspace{2mm}
\noindent{\bf Performance Measures.} Accuracy, precision, recall and F1-score are commonly used measures for binary classification. Following prior works such as~\cite{hmt,DBLP:conf/kdd/0001S19}, we consider our flood segmentation problem as a pixel-wise binary classification problem, and report the above four measures in two contexts: (1)~``Dry'' where dry (resp.\ flood) is the positive (resp.\ negative) class, and (2)~``Flood'' where dry (resp.\ flood) is the positive (resp.\ negative) class. These measures are computed only from labeled pixels in the test images.

\vspace{2mm}
\noindent{\bf Comparison with Baselines.} Here, we consider 3 representative baselines: (1)~HMT~\cite{hmt}, which is a graphical model for transductive learning that considers both the disaster-time RGB image and the elevation map, (2)~U-Net 3C, which is a U-Net model that only considers the disaster-time RGB image, (3)~U-Net 7C, which is a U-Net model that considers both the RGB images and the elevation map. They are compared with EvaNet 7C in Table~\ref{table:model_comparison}. We skip 4C models since they are dominated by 7C ones (c.f.\ Table~\ref{table:model_comparison_4c_7c} in our online appendix~\cite{appendix}).

We can see that EvaNet 7C consistently gives the highest accuracy and F1-score, much higher than all the baselines. Among the other models, HMT performs well on R3, R4 and R7, but very poorly on R5 and R6 where more pixels are predicted as dry, which shows that HMT is region-sensitive and not very robust. U-Net 7C is the second best on R3, R5, and R6 and generally better than U-Net 3C and HMT, but U-Net 7C is even worse than U-Net 3C on R4. We explored the data and found that our pixels in training regions all have elevation below 80 meters, but quite a few pixels in R4 are higher than 90 meters, leading to poor model generalization when the elevation map is treated as an additional input channel backfiring on performance. However, our way of using the elevation map in EvaNet 7C does not cause any generalization issue, and EvaNet 7C still gives the best performance on R4;  it also outperforms all other models on R6 and R7, which is from a different flooding event. This shows that our EvaNet design is robust thanks to the guidance by the physical law of gravity.

\vspace{2mm}
\noindent{\bf Ablation Study.} To verify the effectiveness of both our Eva-Conv and Eva-Reg designs, we conduct an ablation study using 7-channel input, by exploring the performance of 4 models: (1)~U-Net 7C (i.e., with loss $\mathcal{L}_{CE}$), (2)~U-Net with our loss function $\mathcal{L}_{eva}$ (denoted by U-Net 7C + Eva-Reg), (3)~U-Net 7C with our new Eva-Conv operations in both encoder and decoder (denoted by U-Net 7C + Eva-Conv), and (4)~EvaNet 7C, i.e., with both Eva-Conv and Eva-Reg. Without loss of generality, Table~\ref{table:ablation_study} shows the performance results on test region R4, where we can see that (i)~both Eva-Reg and Eva-Conv improve the model accuracy and F1-score as compared with U-Net 7C, (ii)~Eva-Reg (i.e., our new loss $\mathcal{L}_{eva}$) improves the performance more significantly, and (iii)~our EvaNet 7C model combining both designs provides the highest accuracy and F1-score. In our online appendix~\cite{appendix}, we show that our techniques are effective also in other ConvNet architectures using Table~\ref{table:ablation_study_fcn}, and \textbf{we also visually inspect the flood maps from different models in Figure~\ref{ablation_figs} to confirm our conclusions}.

\vspace{1mm}
\noindent{\bf Effect as a U-Net Drop-in Replacement.} We also find that EvaNet can improve the performance of an existing model such as HMCT-PP~\cite{DBLP:journals/tkde/SainjuHJ22} (i.e., U-Net + HMCT) by replacing its U-Net component. Due to space limit, please refer to Table~\ref{table:hmct} in our online appendix~\cite{appendix} for more details.

\section{Conclusion}
We presented EvaNet, a physics-guided segmentation model for flood extent mapping on satellite imagery. EvaNet is based on the popular encoder-decoder architecture with two novel techniques: (1)~an elevation-guided loss function, and (2)~an elevation-regulated convolution. Extensive experiments show that both techniques are effective, and that EvaNet works as a perfect drop-in replacement for U-Net in flood mapping solutions such as HMCT-PP.

\section*{Acknowledgements}
The work was done when Dr.\ Mirza Tanzim Sami and Dr.\ Jalal Khalil worked as PhD students under the supervision of Da Yan at the University of Alabama at Birmingham. The research of Mirza Tanzim Sami, Da Yan, Saugat Adhikari, Lyuheng Yuan and Jiao Han were supported by NSF OAC-2106461, ARDEF 1ARDEF21 03, NSF OAC-2313192, DOE ECRP Award 0000274975, South Big Data Innovation Hub 2022 S.E.E.D.S.\ Award, NSF OAC-2414185, and NSF OAC-2414474. 
Mirza Tanzim Sami acknowledges the financial support from the Alabama Graduate Research Scholars Program (GRSP) funded through the Alabama Commission for Higher Education and administered by the Alabama EPSCoR. 
The research of Zhe Jiang was supported by NSF OAC-2152085, NSF IIS-2147908 and NSF IIS-2207072. 
The research of Yang Zhou was supported by NSF OAC-2313191.

\bibliographystyle{named}
\bibliography{ijcai23}

\clearpage
\appendix
\section{Appendix}

\subsection{Related Work on KGML}
Two works utilize Fourier Neural Operator (FNO) to emulate (but speed up) hydrodynamic modeling/simulation for flood forecasting: \cite{DBLP:conf/iccvw/SunLLHS023} uses CREST-iMAP to generate training data on historical storm events where inputs to FNO include antecedent precipitation, simulated water depths, and DEM; while \cite{jiang2021digital} builds an FNO-based surrogate for NEMO to produce accelerated emulation of coastal dynamics. Alternatives to FNO (to speed up PDE evaluation) for flood forecasting have been studied such as PINN~\cite{mahesh2022physics}. There are also works that use LSTM to capture the temporal process, e.g., for streamflow prediction~\cite{DBLP:conf/kdd/GhoshRTLKJDN022} and rainfall-runoff modeling~\cite{nearing2020deep}.

\subsection{Distribution of Pixel-Pair Cases}

\begin{figure}[h]
	\centering
	\includegraphics[width=\columnwidth]{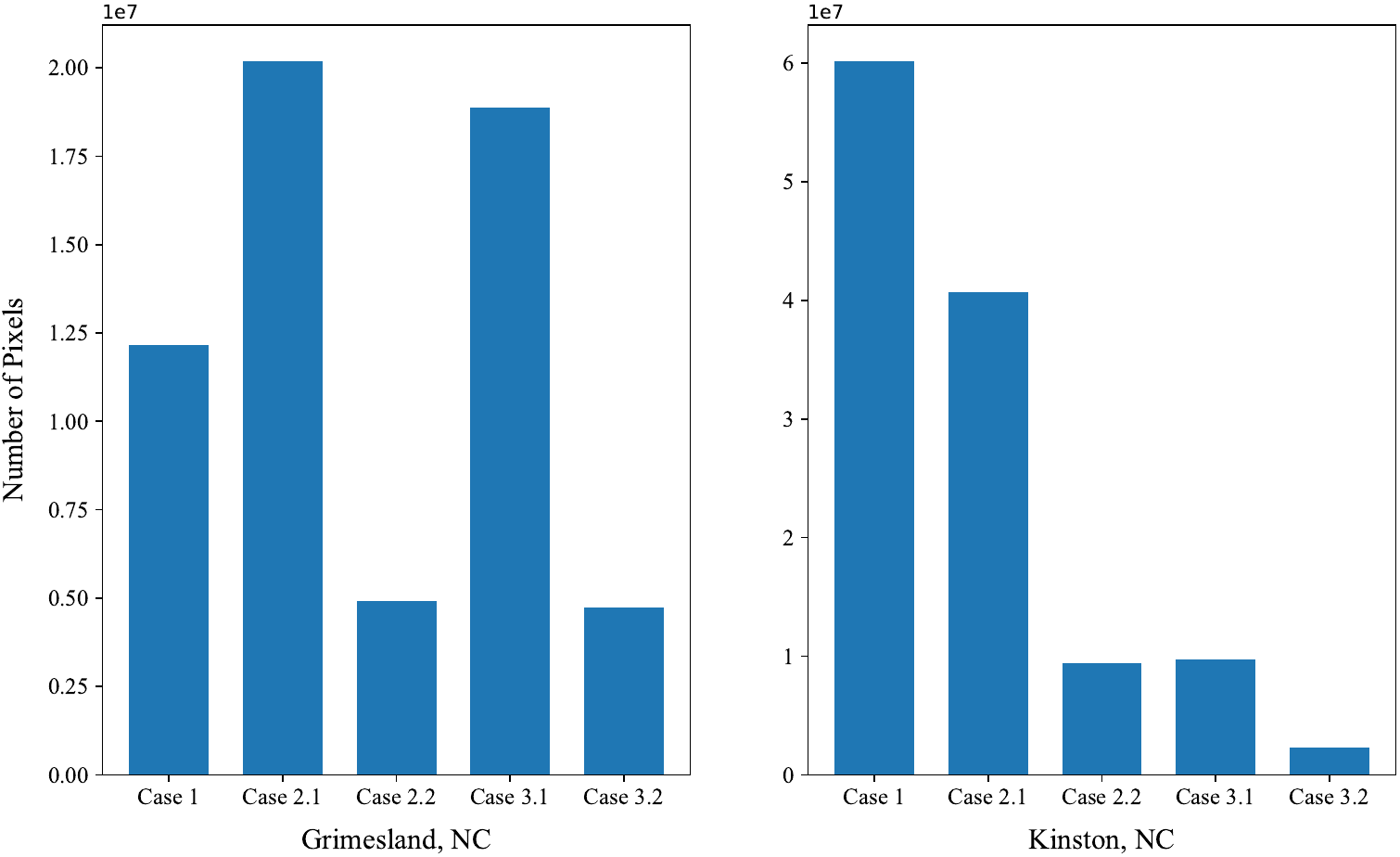}
	\caption{Histograms of the five cases.}\label{Bar_1_3}
\end{figure}

Figure~\ref{Bar_1_3} shows the number of pixels falling in each case on a training region in Grimesland, NC and a test region in Kinston, NC used in our experiments. We see that the test region has a lot of annotated pixels so Case~1 has the most pixels. Other than Case~1, Case~2.1 is the most frequent case in both regions, followed by Case~3.1, so we expect the physical constraints to provide sufficient regularization.

\begin{figure}[t]
	\centering
	\includegraphics[width=\columnwidth]{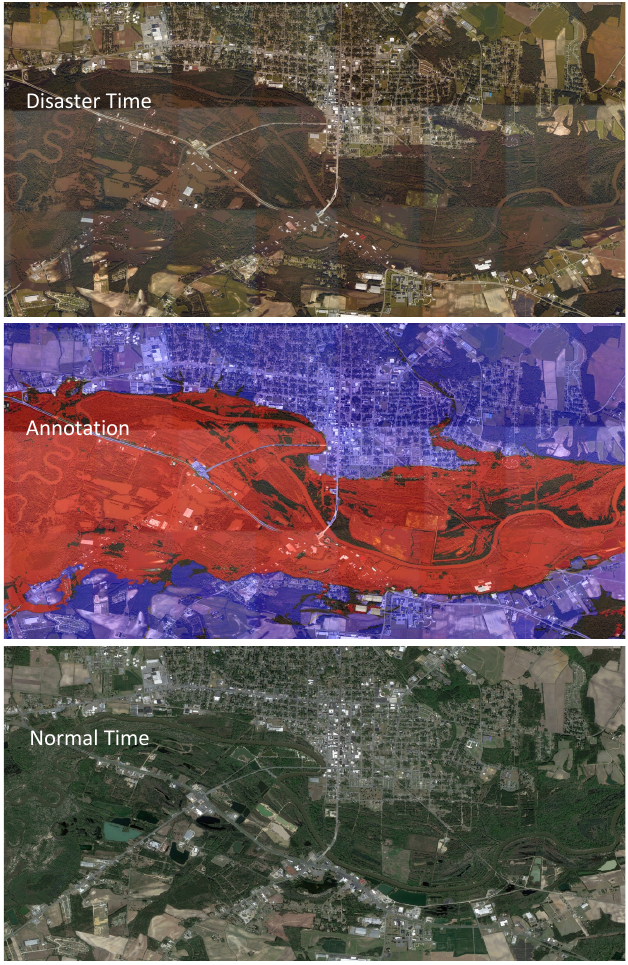}
	\caption{Imagery and annotation of R4: Kinston, NC.}\label{kinston}
\end{figure}


\subsection{Input Data Visualization}
Figure~\ref{kinston} shows the imagery for Kinston, NC during Hurricane Matthew in 2016 (top), and its annotation (middle) where red (resp.\ blue) pixels are labeled as flooded (resp.\ dry). There are some unlabeled pixels covered by tree canopies, which are not included in the loss computation during traing, and not included during test when calculating the evaluation measures. Figure~\ref{kinston} also shows the normal-time imagery (bottom) obtained from Google Earth. We align the normal-time image with the disaster-time one using Georeferencer in QGIS, and rotate them so that they are axis-aligned for ease of patch cutting. We expect that inputting the normal-time image into a model is effective in helping identify flooded regions by providing a contrast.

\subsection{Model Setting Tuning}

\begin{table*}[t]
    \centering
        \begin{tabular}{c || c c c c|c c c c}
        \hline
        \multirow{2}{*}{Method} & \multicolumn{4}{c|}{Dry} & \multicolumn{4}{c}{Flood}\\ 
        \cline{2-9}
        & Accuracy & Precision & Recall & F1-Score  & Accuracy & Precision & Recall & F1-Score  \\   \midrule
        \multicolumn{1}{c||}{GMU}  & 88.78 & 83.90 & 98.31 & 90.54 & 88.78 & 97.44 & 77.34 & 86.24 \\ 
        \multicolumn{1}{c||}{GIF}  &  90.62 & 85.38 & \textbf{99.93} & 92.08  & 90.62  & \textbf{99.90} & 79.44 & 88.50\\  
        \multicolumn{1}{c||}{GLU}  &  \textbf{92.15} & \textbf{88.64} & 98.22 & \textbf{93.18}  & \textbf{92.16}  & 97.54 & \textbf{84.89} & \textbf{90.77}\\  
        \bottomrule
        \end{tabular}%
        \caption{Fusion techniques (unit: \%) on R4.}
        \label{table:fusiopn_comparison_4c_7c}
\end{table*}

\begin{table*}[t]
 \centering
 \begin{tabular}{c|| c c c c|c c c c}
 \toprule
\multirow{2}{*}{Method} & \multicolumn{4}{c|}{Dry} & \multicolumn{4}{c}{Flood}\\ 
 \cline{2-9}
 & Accuracy & Precision & Recall & F1-Score  & Accuracy & Precision & Recall & F1-Score  \\ 
   \midrule
  \multicolumn{1}{c||}{CE}  &  83.50 & 77.73 & 97.79 & 86.61 & 83.50 & 96.15 & 66.34 & 78.51\\ 
 \multicolumn{1}{c||}{CE + Eva-Reg}  & 89.92 & 84.48 & \textbf{99.88} & 91.54 & 89.92 & \textbf{99.81} & 77.96 & 87.55  \\ 
 \multicolumn{1}{c||}{Eva-Reg}  & \textbf{92.16} & \textbf{88.64} & 98.22 & \textbf{93.18} & \textbf{92.16} & 97.54 & \textbf{84.88} & \textbf{90.77} \\   
  \bottomrule
  \end{tabular}
   \caption{Loss Comparison (Unit: \%) on Test Region R4: Kinston, NC}
   \label{table:loss_comparison}
\end{table*}

\begin{table*}[t]
 \centering
 \begin{tabular}{c|| c c c c|c c c c}
 \toprule
\multirow{2}{*}{Method} & \multicolumn{4}{c|}{Dry} & \multicolumn{4}{c}{Flood}\\ 
 \cline{2-9}
 & Accuracy & Precision & Recall & F1-Score  & Accuracy & Precision & Recall & F1-Score  \\ 
   \midrule
 \multicolumn{1}{c}{} & \multicolumn{8}{c}{Falkland, NC}\\
 \midrule
 \multicolumn{1}{c||}{Eva-Diff Weighting}  & 95.97 & 98.45 & 95.63 & 97.02 & 95.97 & 90.97 & 96.69 & 93.75  \\ 
 \multicolumn{1}{c||}{Log Eva-Diff Weighting}  & 95.92 & \textbf{98.58} & 95.45 & 96.99 & 95.92 & 90.64 & \textbf{96.97} & 93.70 \\ 
 \multicolumn{1}{c||}{Binary Weighting}  & \textbf{96.14} & 98.13 & \textbf{96.22} & \textbf{97.16} & \textbf{96.14} & \textbf{92.02} & 95.96 & \textbf{93.95} \\  
   \midrule
 \multicolumn{1}{c}{}   & \multicolumn{8}{c}{Kinston, NC}\\
    \midrule
 \multicolumn{1}{c||}{Eva-Diff Weighting}  & 80.59 & 74.88 & 96.96 & 84.50 & 80.59 & 94.34 & 60.92 & 74.03  \\ 
 \multicolumn{1}{c||}{Log Eva-Diff Weighting}  & 81.74 & 75.10 & \textbf{99.56} & 85.61 & 81.74 & \textbf{99.13} & 60.34 & 75.01 \\ 
 \multicolumn{1}{c||}{Binary Weighting}  & \textbf{92.16} & \textbf{88.64} & 98.22 & \textbf{93.18} & \textbf{92.16} & 97.54 & \textbf{84.88} & \textbf{90.77} \\ 
   \midrule
   \multicolumn{1}{c}{}   & \multicolumn{8}{c}{Greenville-West, NC}\\
    \midrule
 \multicolumn{1}{c||}{Eva-Diff Weighting}  & 93.23 & 96.28 & 92.67 & 94.44 & 93.22 & 88.70 & 94.15 & 91.34  \\ 
 \multicolumn{1}{c||}{Log Eva-Diff Weighting}  & 94.33 & 97.90 & 92.86 & 95.31 & 94.33 & 89.23 & 96.75 & 92.84 \\ 
 \multicolumn{1}{c||}{Binary Weighting}  &  \textbf{97.56} & \textbf{98.13} & \textbf{97.94} & \textbf{98.04} & \textbf{97.56} & \textbf{96.64} & \textbf{96.95} & \textbf{96.80}\\ 
  \bottomrule
  \end{tabular}
   \caption{Weighting Scheme Comparison (Unit: \%)}
   \label{table:weighting_scheme_comparison}
\end{table*}

\vspace{2mm}
\noindent{\bf Effect of Fusion Operations.} Recall from Figure~\ref{arch} that we use 8 fusion operations (that generate the fused purple tensors). Table~\ref{table:fusiopn_comparison_4c_7c} now shows the performance of our model on test region R4, as well as its two variants where the 8 GLU operations are replaced with alternative fusion operations GIF~\cite{arevalo2017gated} and GMU~\cite{kim2018robust}, respectively. We do not observe clear advantages of applying GIF and GMU over GLU in Figure~\ref{arch}, and the results are similar for the other test regions. Note that GLU is now widely adopted in recent works such as~\cite{DBLP:conf/ijcai/WuPLJZ19,DBLP:conf/kdd/Hui0CK21,DBLP:conf/icdm/Hui0CK21}, rather than other fusion methods like GIF and GMU.

\vspace{2mm}
\noindent{\bf Effect of Loss Functions.}
Recall that our EvaNet uses $\mathcal{L}_{eva}$ as the loss function by default. This is because we find that it performs the best in our experiments. For example, Table~\ref{table:loss_comparison} compare the performance of three loss functions for EvaNet 7C on region R4: (1)~$\mathcal{L}_{CE}$, (2)~$\mathcal{L}_{CE}+\mathcal{L}_{eva}$, and (3)~$\mathcal{L}_{eva}$. We can see that $\mathcal{L}_{eva}$ outperforms the other loss functions in both accuracy and F1-score, which shows that $\mathcal{L}_{eva}$ is an excellent loss function on its own.

\vspace{2mm}
\noindent{\bf Effect of Weighting Scheme.}
Recall from Eq~(2) that $\mathcal{L}_{eva}$ uses a weight term $w(\mathbf{p}, \mathbf{p}_n)$ for which we proposed three schemes: elevation-difference weighting in Eq~(5), log-elevation-difference weighting in Eq~(6), and binary weighting in Eq~(4). Table~\ref{table:weighting_scheme_comparison} compares our EvaNet 7C variants where $w(\mathbf{p}, \mathbf{p}_n)$ uses these three schemes, and we can see that the default scheme `binary weighting' is the clear winner in terms of both accuracy and F1-score.

\begin{table*}[t]
 \centering
 \begin{tabular}{c | c || c c c c|c c c c}
 \toprule
\multirow{2}{*}{Elev Path} & \multirow{2}{*}{Img Path} & \multicolumn{4}{c|}{Dry} & \multicolumn{4}{c}{Flood}\\ 
 \cline{3-10}
& & Accuracy & Precision & Recall & F1-Score  & Accuracy & Precision & Recall & F1-Score  \\
 \midrule
 Max Pool & \multicolumn{1}{c||}{Max Pool}  &  89.70 & 86.39 & 96.30 & 91.08 & 89.70 & 94.84 & 81.78 & 87.83\\ 
 Avg Pool & \multicolumn{1}{c||}{Avg Pool}  &  89.88 & 87.96 & 94.38 & 91.06 & 89.88 & 92.60 & 84.48 & 88.36\\  
 Max Pool & \multicolumn{1}{c||}{Avg Pool}  &  85.55 & 80.84 & 96.36 & 87.92 & 85.55 & 94.32 & 72.56 & 82.02 \\ 
 Avg Pool & \multicolumn{1}{c||}{Max Pool}  &  \textbf{97.56} & \textbf{98.13} & \textbf{97.94} & \textbf{98.04} & \textbf{97.56} & \textbf{96.64} & \textbf{96.95} & \textbf{96.80}\\
  \bottomrule
  \end{tabular}
   \caption{Pooling Scheme Comparison (Unit: \%) on Test Region R4: Kinston, NC}
   \label{table:pooling_scheme_comparison}
\end{table*}

\vspace{2mm}
\noindent{\bf Effect of Pooling Scheme.}
Recall from Figure~\ref{arch} that by default, EvaNet's encoder uses max pooling for the spectral path (Img Path), and average pooling for the elevation path (Elev Path). We now compare the various pooling options for both paths. Table~\ref{table:pooling_scheme_comparison} shows the comparison results on test region R4, where we can see that our default setting provides the best performance in all measures.

\begin{table*}[t]
 \centering
 \begin{tabular}{c|| c c c c|c c c c}
 \toprule
\multirow{2}{*}{Method} & \multicolumn{4}{c|}{Dry} & \multicolumn{4}{c}{Flood}\\ 
 \cline{2-9}
 & Accuracy & Precision & Recall & F1-Score  & Accuracy & Precision & Recall & F1-Score  \\ 
 \midrule
 \multicolumn{1}{c||}{2 Blocks}  & 89.05 & 83.75 & \textbf{99.18} & 90.81 & 89.05 & \textbf{98.74} & 76.88 & 86.44 \\ 
 \multicolumn{1}{c||}{3 Blocks}  & \textbf{92.16} & \textbf{88.64} & 98.21 & \textbf{93.18} & \textbf{92.16} & 97.54 & \textbf{84.88} & \textbf{90.77}\\ 
 \multicolumn{1}{c||}{4 Blocks}  & 86.86 & 81.08 & 99.02 & 89.16 & 86.86 & 98.40 & 72.25 & 83.32\\  
 \multicolumn{1}{c||}{5 Blocks}  & 86.69 & 81.36 & 98.08 & 88.94 & 86.69 & 96.93 & 73.01 & 83.29\\ 
  \bottomrule
  \end{tabular}
   \caption{Effect of Blocks (Unit: \%) on Test Region R4: Kinston, NC}
   \label{table:effect_of_blocks}
\end{table*}

\vspace{2mm}
\noindent{\bf Effect of Number of Blocks.}
Recall from Figure~\ref{arch} that our encoder and decoder use 3 convolution blocks. We now explore the effect if we change the number of blocks. For example, in the 4-block setting, the bottleneck layer is further downsampled into an $8\times8\times64$ tensor. Without loss of generality, Table~\ref{table:effect_of_blocks} compares our EvaNet 7C variants with 2, 3, 4 and 5 blocks when R4 is used for testing. We can see that the accuracy and F1-score peak in the 3-block setting, and then drop slightly as the number of blocks increases further.

\begin{table*}[t]
    \centering
        \fontsize{9}{11}
        \begin{tabular}{c || c c c c|c c c c}
        \hline
        \multirow{2}{*}{Method} & \multicolumn{4}{c|}{Dry} & \multicolumn{4}{c}{Flood}\\ 
        \cline{2-9}
        & Accuracy & Precision & Recall & F1-Score  & Accuracy & Precision & Recall & F1-Score  \\   \midrule
        \multicolumn{1}{c||}{U-Net 4C}  & 74.99 & 69.14 & \textbf{97.87} & 81.03 & 74.99 & 94.89 & 47.51 & 63.32 \\ 
        \multicolumn{1}{c||}{U-Net 7C}  &  \textbf{83.90} & \textbf{78.45} & 97.20 & \textbf{86.82}  & \textbf{83.90}  & \textbf{95.28} & \textbf{67.92} & \textbf{79.31}\\  
        \midrule
        \multicolumn{1}{c||}{EvaNet 4C}  & 90.14 & 85.65 & \textbf{98.42} & 91.59 & 90.14 & \textbf{97.69} & 80.19 & 88.08 \\
        \multicolumn{1}{c||}{EvaNet 7C}  & \textbf{92.16}  & \textbf{88.64} & 98.22 & \textbf{93.18}  & \textbf{92.16} & 97.54 & \textbf{84.88} & \textbf{90.77} \\ 
        \bottomrule
        \end{tabular}%
        \caption{4 Channels v.s.\ 7 Channels (Unit: \%) on Test Region R4: Kinston, NC}
        \label{table:model_comparison_4c_7c}
\end{table*}

\begin{figure*}[t]
	\centering
	\includegraphics[width=1.8\columnwidth]{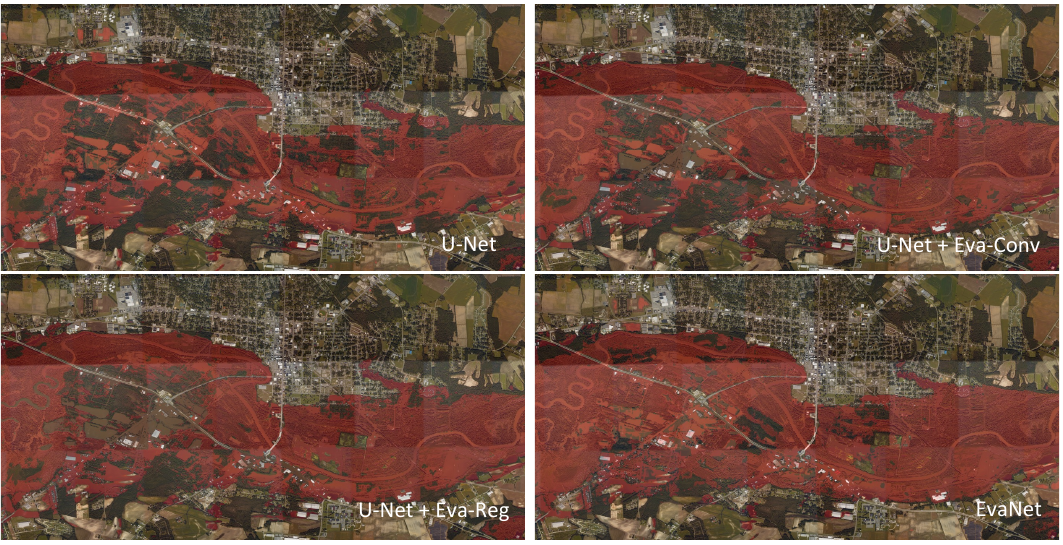}
	\caption{Visualization of model predictions.}\label{ablation_figs}
\end{figure*}

\subsection{Effectiveness of Normal-Time Imagery} Here we answer the following question: {\em whether adding the RGB channels of the normal-time image improves performance compared with using disaster-time image plus elevation map only?}

The answer is affirmative. For example, Table~\ref{table:model_comparison_4c_7c} shows the performance of U-Net 4C and 7C (resp.\ EvaNet 4C and 7C), and we can see that the 7C models significantly outperforms 4C ones in terms of both accuracy and F1-score.

\begin{table*}[t]
 \centering
 \begin{tabular}{c || c c c c|c c c c}
    \toprule
    \multirow{2}{*}{Method} & \multicolumn{4}{c|}{Dry} & \multicolumn{4}{c}{Flood}\\ 
     \cline{2-9}
     & Accuracy & Precision & Recall & F1-Score  & Accuracy & Precision & Recall & F1-Score  \\ 
     \midrule
     \multicolumn{1}{c||}{FCN8s 7C}  & 76.76 & 70.26 & \textbf{99.52} & 82.37 & 76.76 & \textbf{98.86} & 49.40 & 65.88  \\ 
     \multicolumn{1}{c||}{FCN8s 7C w/ Eva-Conv}  & 82.89 & 77.27 & 97.24 & 86.11 & 82.89 & 95.20 & 65.63 & 77.70  \\  
     \multicolumn{1}{c||}{FCN8s 7C w/ Eva-Reg}  & 86.70 & 82.00 & 96.89 & 88.83 & 86.70 & 95.23 & 74.46 & 83.58   \\ 
     \multicolumn{1}{c||}{Eva-FCN8s 7C}  & \textbf{87.84} & \textbf{85.48} & 93.62 & \textbf{89.37} & \textbf{87.84} & 91.35 & \textbf{80.90} & \textbf{85.80}  \\ 
      \bottomrule
      \end{tabular}
   \caption{Ablation study (unit: \%) on test region R4.}
   \label{table:ablation_study_fcn}
\end{table*}

\subsection{Ablation Study with FCN-8s} Recall from Table~\ref{table:ablation_study} that when U-Net is used as the encoder-decoder backbone, both Eva-Reg and Eva-Conv improve the model performance. We remark that this improvement carries over to other ConvNet architectures. For example, Table~\ref{table:ablation_study_fcn} shows the ablation study results when the encoder-decoder backbone is FCN-8s from~\cite{long2015fully} instead of U-Net, where we observe similar conclusions.

Figure~\ref{ablation_figs} shows our flood prediction results (pixels with flood probability $\geq50\%$ are masked in red), where we can see that only EvaNet 7C can fill the entire flooded areas.

\begin{table*}[t]
 \centering
 \begin{tabular}{c|| c c c c|c c c c}
     \toprule
    \multirow{2}{*}{Method} & \multicolumn{4}{c|}{Dry} & \multicolumn{4}{c}{Flood}\\ 
     \cline{2-9}
     & Accuracy & Precision & Recall & F1-Score  & Accuracy & Precision & Recall & F1-Score  \\ 
     \midrule
     \multicolumn{1}{c||}{U-Net 7C}  & 83.90  & 78.45 & 97.20 & 86.82 & 83.90  & 95.28 & 67.92 & 79.31\\ 
     \multicolumn{1}{c||}{U-Net 7C + HMCT}  & 93.71  & 91.31 & 97.78 & 94.44 & 93.71  & 97.09 & 88.82 & 92.77\\ 
     \multicolumn{1}{c||}{EvaNet 7C}  & 92.16  & 88.64 & \textbf{98.21} & 93.18 & 92.16  & \textbf{97.54} & 84.88 & 90.77 \\  
     \multicolumn{1}{c||}{EvaNet 7C + HMCT}  & \textbf{94.38}  & \textbf{97.16} & 92.40 & \textbf{94.72}  & \textbf{94.38}  & 91.38 & \textbf{96.75} & \textbf{93.99}\\ 
      \bottomrule
      \end{tabular}
   \caption{Effect of HMCT postprocessor (unit: \%) on R4.}
   \label{table:hmct}
\end{table*}

\subsection{Effect as a U-Net Drop-in Replacement} Here we answer the following question: {\em can EvaNet improve the performance of an existing model by replacing its U-Net component?} 
Recall from Section~\ref{sec:related} that the HMCT-PP  (i.e., U-Net + HMCT) model~\cite{DBLP:journals/tkde/SainjuHJ22} acts as a postprocessor of U-Net to further improve performance. If we replace U-Net with EvaNet, we hope that the HMCT postprocessor will provide an even better performance. Our experiments provides an affirmative answer. For example, Table~\ref{table:hmct} shows the performance of U-Net 7C and EvaNet 7C on test region R4, w/ and w/o HMCT postprocessing, respectively. We can see that `EvaNet 7C + HMCT' gives the best performance, outperforming `U-Net 7C + HMCT.'

\end{document}